\documentclass{llncs}

\usepackage[utf8]{inputenc}
\usepackage[T1]{fontenc}
\usepackage[english]{babel}
\usepackage{csquotes}
\usepackage{amsmath}
\usepackage{amssymb}
\usepackage{booktabs}
\usepackage{graphicx}
\usepackage{hyperref}
\usepackage{listings}
\usepackage{algorithm}
\usepackage[noend]{algpseudocode}
\usepackage{wrapfig}
\usepackage{siunitx}
 \usepackage[
    backend=biber,
    style=lncs,
    maxbibnames=3,
  ]{biblatex}

\begin{document}

\title{Bottom-up Anytime Discovery of Generalised Multimodal Graph Patterns for Knowledge Graphs
    \thanks{This research was funded by CLARIAH-PLUS (NWO Grant 184.034.023)}}

    \author{Wilcke WX\inst{1}\orcidID{0000-0003-2415-8438} \and
        Mourits RJ\inst{2}\orcidID{0000-0002-2267-1679} \and\\
        Rijpma, A\inst{3}\orcidID{0000-0002-8950-8227} \and
        Zijdeman RL\inst{2,4}\orcidID{0000-0003-3902-3720}}

\authorrunning{Wilcke WX, et al.}
\institute{Dept.\ of Computer Science, Vrije Universiteit Amsterdam, The Netherlands
    \and Data \& Augmentation, International Institute for Social History, The Netherlands
    \and Economic and Social History, Utrecht University, The Netherlands
    \and University of Stirling, Scotland UK\\\vspace{1mm}
\email{w.x.wilcke@vu.nl, rick.mourits@iisg.knaw.nl\\a.rijpma@uu.nl, richard.zijdeman@iisg.knaw.nl}}

\maketitle

\begin{abstract} 
    Vast amounts of heterogeneous knowledge are becoming publicly available in the form of knowledge graphs,
    often linking multiple sources of data that have never been together before, and thereby enabling scholars to answer
    many new research questions. It is often not known beforehand, however, which questions the data might have the answers
    to, potentially leaving many interesting and novel insights to remain undiscovered. To support scholars during
    this scientific workflow, we introduce an \textit{anytime} algorithm for the bottom-up discovery of generalized
    multimodal graph patterns in knowledge graphs. Each pattern is a conjunction of binary statements with (data-) type
    variables, constants, and/or value patterns. Upon discovery, the patterns are converted to SPARQL queries and presented
    in an interactive facet browser together with metadata and provenance information, enabling scholars to explore,
    analyse, and share queries. We evaluate our method from a user perspective, with the help of domain
    experts in the humanities.

    \keywords{Pattern Mining \and Hypothesis Generation \and Heterogeneous Knowledge \and Generalized Graph Patterns \and Knowledge Graphs}
\end{abstract}

\section{Introduction}

In only a short span of time, knowledge graphs have transitioned from an academic curiosity to an attractive data model
for storing and publishing scientific data~\cite{DBLP:series/synthesis/2021Hogan}. Amongst the multitude of adopters of
this data model are
many of the world's galleries, libraries, archival institutions, and
museums~\cite{DBLP:conf/esws/BoerWGHIOS12,DBLP:reference/bdt/HaslhoferIS19,DBLP:conf/esws/SzekelyKYZFAG13}, as
well as various scientific communities including linguistics, archaeology, humanities, and
history~\cite{DBLP:conf/lrec/DeclerckMHGCMCR20,DBLP:journals/jocch/MeghiniSRWGCFFH17,DBLP:series/ssw/Merono-PenuelaB20}.
The combined efforts of these institutes and communities have resulted in a considerable number of publicly-available
knowledge graphs which, together, surmount to vast amounts of interconnected heterogeneous knowledge. Much of this
knowledge used to be stored in analogue or digital silos, and has never been brought together before. Now linked to one
another, this federative network of knowledge offers great opportunities for scholars, who can now potentially ask and
answer many new research questions.

Drafting research questions is an essential step in the scientific research workflow. Such questions can either be
derived from the scholarly literature or from patterns in data. However, without study, it is often not known beforehand
which questions these data might have the answers to. Even if accompanied by rich metadata, these alone are often not
enough to guide scholars in this process, limiting them to insights from the literature or sparks of their own
imagination. This may result in many potentially interesting and novel insights to remain unstudied, due to possible
biases or blind spots in the literature and the scholars' thinking. This work aims to support scholars during this
early stage of the scientific workflow, by highlighting potentially interesting patterns in their data that may form the
building blocks for new research questions, and which can be used as evidence for already existing lines of research.

Pattern detection on graph-shaped data can take on various forms. On the most fundamental level, graph patterns are
recurrent and statistically significant subgraphs in which some or all of the vertices have been replaced by unbound
variables~\cite{DBLP:conf/icdm/KuramochiK01}. Generalized graph patterns take this a step further, by having special
variables that cover an entire set of vertices, such as all members of a certain class~\cite{DBLP:conf/icdm/Inokuchi04}.
Scholars can use such patterns to explore \emph{structural} regularities in the graph; other regularities, such as those
between the various numerical, temporal, and textual attributes values are generally not considered, however, despite
their prevalence in many knowledge graphs. This is particularly evident in the soft sciences where measurements, dating,
and note taking are commonplace~\cite{schoch2013big}. Since these multimodal data often contain insightful and unique
information about the subject they belong to, it becomes all the more important to treat them as first-class citizen. By
doing so, we can integrate non-structural regularities into generalized graph patterns and obtain more expressive
patterns that offers scholars a more fine-grained view of their data.

\begin{figure}[t]
    \label{fig:example_graph}
    \includegraphics[width=\textwidth]{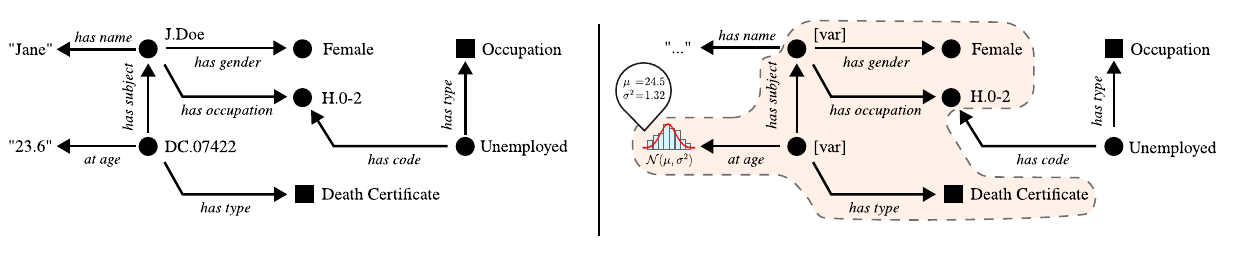}
    \caption{An example of a subgraph in the civil registry domain (left) and a possible graph pattern (right).
        Circles and squares represent entities and classes, respectively, and attribute values are within quotation
        marks. The coloured area indicates the structural component of the pattern, whereas the distribution conveys the
        non-structural component.}
\end{figure}

Figure~\ref{fig:example_graph} illustrates the merit of combining structural and non-structural regularities. The left
side of the figure depicts a civil record about an unemployed woman, named Jane, who died at an age of 23.6 years old,
whereas, on the right, a graph pattern is shown that covers this record. The depicted pattern likewise covers other
records about unemployed women, provided that they died at an age that falls within the learned
distribution\footnotemark. This distribution is an example of a non-structural regularity; without it, the pattern would
have been limited to unemployed women whose age of death is on record, irrespective of the value. For other attributes,
in this example the person's name, the variation between values might be too great to constitute a regularity, hence it
being excluded\ from the pattern. 

\footnotetext{We maintain $\mu \pm \sigma$ as the range for a match.}

In this work, we introduce an algorithm for the \textit{bottom-up} discovery of generalized multimodal graph patterns in
knowledge graphs. The patterns are generated directly from the graph, leveraging both statistics and semantics to guide
the discovery process, and get more precise with each following iteration. This gives our algorithm the \textit{anytime}
property, as scholars can terminate the process at their leisure and still obtain potentially interesting, albeit more
generic, patterns. Additionally, special attention is given to the multimodal nature of many knowledge graphs, by
allowing the combination of structural regularities with those between numerical, temporal, and textual attributes.
Moreover, to mitigate the curse of dimensionality, our algorithm incorporates smart pruning strategies and other
optimization techniques.

We evaluate our method and the patterns it yields from a user perspective, by asking feedback from the target
community via a focussed questionnaire. To lessen the semantic gap, the patterns are automatically converted to SPARQL
queries upon discovery, improving their interpretability and familiarity. As a final step, the queries are presented in
an interactive facet browser together with metadata, provenance information, and graph visualizations, enabling scholars
to explore and analyse the patterns, as well as reproduce and share relevant data selections. 

To summarize, our main contributions are a) a novel \textit{anytime} algorithm for the \textit{bottom-up} discovery of
generalized multimodal graph patterns in knowledge graphs, b) the natural integration of various non-structural
regularities into generalized graph patterns, and c) an extensive evaluation together with domain experts.

\section{Related Work}

Pattern detection methods generally fall in one of two categories: the symbolic approaches, which employ some form of
(logical) rule mining, and the non-symbolic approaches, which often involve (graph) neural networks in an unsupervised
learning setting. The method described in this paper falls into the first category.

\subsubsection{Rule Mining for Graphs}
A considerable body of literature is available on rule mining for relation data. Many of these approach the problem from
either a \textit{frequentist} or \textit{inductive} school of thought. Methods that belong to the latter
(e.g.~\cite{DBLP:journals/ngc/Muggleton95,DBLP:conf/alt/MuggletonF90,DBLP:journals/ml/Quinlan90}) generally apply
\textit{inductive logic programming}, which involves learning logical rules that explain all true arcs and no false
ones~\cite{DBLP:journals/ngc/Muggleton91}. This technique is less suited for knowledge graphs, however, due to
challenges with scalability and the need for a closed world~\cite{DBLP:journals/ml/MuggletonRPBFIS12}. 
In contrast, frequentist approaches emphasize (relative) coverage, and
typically involves the mining of association rules or frequent substructures. This work falls under the umbrella of
frequentist approaches, by introducing a method to discover statistically relevant substructures in knowledge graphs.

Frequent subgraph mining is a mature field of research which involves finding all subgraphs that occur more than
some predefined number of times~\cite{DBLP:journals/ker/JiangCZ13}. These subgraphs, and the graphs from which they are
mined, are assumed to be labelled simple graphs. For knowledge graphs, in which the vertices and arcs have types, it
is therefore more interesting to look for \textit{generalized} subgraphs, for example by abstracting away to the level of the
classes~\cite{DBLP:journals/kbs/BaratiBL17,DBLP:conf/sac/MartinFVDL22,DBLP:conf/pkdd/PalmeW22}, or by generalizing over controlled
vocabularies and taxonomies~\cite{DBLP:conf/edbt/CakmakO08,DBLP:conf/icdim/PetermannMBPR17}. To discover the subgraphs,
these methods commonly employ a bottom-up approach, similar to our method, that begins with the most basic rules and
incrementally adds new arcss until some condition is reached or the search space has been exhausted.

Closely related to subgraph discovery is graph-based association rule mining, which aims at finding rules that
imply frequent co-occurrences between subgraphs. Such rules can be found by adapting the Apriori algorithm for
graph data, in which case so-called \textit{item sets} of correlated arcs are sought, which are then clustered
hierarchically to induce an ordering on frequency~\cite{DBLP:conf/pricai/BaratiBL16,ramezani2014swapriori,DBLP:journals/ws/WilckeBKHS19}.
Other methods are specifically tailored to graphs and use a bottom-up
approach comparable to many subgraph mining methods~\cite{DBLP:conf/www/GalarragaTHS13,DBLP:conf/ki/MeilickeCRS19,DBLP:conf/ic3k/WilckeKBSH20}. In some cases,
background knowledge is levered to infer implicit knowledge~\cite{miani2009narfo,DBLP:journals/kbs/NebotL12}.
 
Many methods employ optimization and smart pruning strategies to reduce the search space to a more manageable
size, for example by cutting unviable branches at an early stage or by avoiding duplicate subgraphs found via
different paths~\cite{DBLP:journals/vldb/GalarragaTHS15,DBLP:conf/esws/LajusGS20,DBLP:conf/ic3k/WilckeKBSH20}. Similar
strategies are being used by our method.

\subsubsection{Query Generation}

To the best of our knowledge, generating queries directly from the data has received little attention. Instead, most literature explores query log
mining~\cite{DBLP:conf/www/ZhangN06}, top-down query construction~\cite{DBLP:conf/acl/YavuzGSY18}, or the use of
language models~\cite{DBLP:conf/sigir/WangSKZ23} for this purpose. A notable exception is
\citeauthor{DBLP:conf/bibm/ShenLSLL15}~\cite{DBLP:conf/bibm/ShenLSLL15}, who cluster biomedical data on semantic
closeness of the relationships and convert these clusters into SPARQL queries. Some studies have also looked into the 
conversion of SPARQL queries into other formats, including logical
formulae~\cite{DBLP:conf/www/Polleres07,DBLP:conf/ki/Schenk07}. Our approach
performs a similar transformation, but in the other direction: from formulae to queries.

\section{Perquisites}

Central to our approach are knowledge graphs and SPARQL queries. This next section will briefly
introduce these concepts.

\subsection{Knowledge Graphs}

A knowledge graph $G = (\mathcal{R}, \mathcal{P}, \mathcal{A})$ is a labelled multidigraph with $\mathcal{R}$ and
$\mathcal{P}$ denoting the set of resources (vertices) and predicates (arc types), respectively, and with $\mathcal{A}
\subseteq \mathcal{P} \times \mathcal{E} \times \mathcal{E} \cup \mathcal{P} \times \mathcal{E} \times \mathcal{L}$
representing the set of all assertions (arcs) that make up the graph. The set of resources $\mathcal{R} = \mathcal{E} \cup
\mathcal{L}$ can be further divided into the set of entities, $\mathcal{E}$, which represent unique things, tangible or
otherwise, and the set of literals, $\mathcal{L}$, which represent attribute values such as text and numbers, and which
belong to exactly one entity.  Literals can optionally be annotated with their datatype (or language tag, from which the
datatype can be inferred) which itself is an entity. 

An example of a knowledge graph is depicted in Figure~\ref{fig:example_graph}-left, showing a small graph from
the civil registry domain. This particular graph contains seven entities, two of which are
classes, and two literals: a number and a string. These elements are linked to each other by exactly eight assertions,
two of which represent the same predicate: \textit{has\_type}.

There are various data models available to model knowledge graphs with. In this work, we consider the \emph{Resource
Description Framework} (RDF)\footnotemark, which is a popular choice for this purpose. However, our approach can be
adapted to other data models with minor changes.

\footnotetext{The RDF specification is available at \url{www.w3.org/TR/rdf11-concepts}}

\subsection{SPARQL Queries}

SPARQL\footnotemark is a query language for RDF-encoded knowledge graphs that supports searching for graph patterns. A
typical SPARQL query consists of three parts: 1) a prologue, in which the namespaces are defined, 2) a \texttt{SELECT}
clause, which specifies the return variables, and 3) a \texttt{WHERE} clause, which contains the graph pattern we are to
match against. SPARQL also provides many other capabilities, but these are out of the scope of this paper.

\footnotetext{The SPARQL specification is available at \url{www.w3.org/TR/sparql11-query}}

Graph patterns in a SPARQL query are similar to their logical counterpart except that the clauses are written in infix
notation---$\mathcal{R} \times \mathcal{P} \times \mathcal{R}$---and that the conjunctions between them are implicit.
Additionally, variables are prepended by a question mark (\texttt{?}), and the \texttt{FILTER} keyword can be used to
constrain the result set. An example is listed in Listing~1-right, showing the SPARQL query corresponding to the graph
pattern in Figure~\ref{fig:example_graph}-right.

\section{Defining Generalized Multimodal Graph Patterns}

Generalized multimodal graph patterns are recurrent and statistically significant subgraphs in which some or all of the
resources have been replaced by special variables. This allows for graph patterns that abstract away from the level of the
individual resources by modelling structural regularities between and non-structural regularities within sets of resources.
From now on, we will refer to such patterns as graph patterns or simply as patterns unless the meaning is not evident from
the context.

Formally, a graph pattern $\phi = c_i \land c_j \land \ldots c_k$ is a conjunction of $k$ clauses, with $k \ge 1$, where
each clause $c = p(a, b)$ is a binary predicate that represents the relationship $p \in \mathcal{P}$ between the elements $a$ and $b$.
Here, $a$ and $b$ can be constants that represent actual resources in the graph, in which case $p(a, b)$ corresponds
to an assertion in $\mathcal{A}$, or they can be variables, representing a set of entities or literals.
In either case, we will refer to $a$ and $b$ as the \emph{head} and \emph{tail} of a relationship, respectively.

In this work we consider three different kinds of variables, namely \textit{object-type}, \textit{data-type}, and
\textit{value-range} variables. The set of resources that each variable
covers is called its \emph{domain}. For each of the three variable types, we define their domain as follows.

\begin{description}
    \itshape
    \item[Object-type:]
    Let $\mathcal{T}_\mathcal{E}$ be the set of object types in $G$, and $T(e, t)$ a binary predicate that holds if
    entity $e \in \mathcal{E}$ is of type $t$. The domain of an object-type variable of type $t \in
    \mathcal{T}_\mathcal{E}$ can now be defined as the set of entities $\mathcal{E}_t \subseteq \mathcal{E}$ such that
    $\forall e \in \mathcal{E}_t: T(e, t)$.

    \item[Data-type:]
    Let $\mathcal{T}_\mathcal{L}$ be the set of datatypes in $G$, and $T(\ell, t)$ a binary predicate that holds if
    literal $\ell \in \mathcal{L}$ is of datatype $t$. The domain of a data-type variable of type $t \in
    \mathcal{T}_\mathcal{L}$ can now be defined as the set of literals $\mathcal{L}_t \subseteq \mathcal{L}$ such that
    $\forall \ell \in \mathcal{L}_t: T(\ell, t)$.

    \item[Value-range:]
    Let predicate $p \in P$ represent a relationship with value space $\mathcal{S} \subseteq \mathcal{L}$ such
    that $\forall \ell \in \mathcal{L}, \exists e \in \mathcal{E}: p(e, \ell) \implies \ell \in \mathcal{S}$.  The
    domain of a value-range variable can now be defined as the set of attribute values $\mathcal{S}_F \subseteq
    \mathcal{S}$ that fall within a distribution $F$ defined on $\mathcal{S}$.
\end{description}

Both object-type and data-type variables allow for a generalization over structure. Examples of the former are the
object types \texttt{Person} and \texttt{Occupation}, which cover all people and jobs, whereas the datatypes
\texttt{String} and \texttt{Float} encompass all text and real-valued attribute values. Value-range variables offer a
further generalization over attribute values, for example by fitting one or more Gaussian distributions on a
collection of years, or by defining a uniform distribution over a set of characters (encoded as regular expression, e.g.
\texttt{"\^{}[:alnum:]\{3,6\}\$"}). 

The clauses in a pattern are subject to several rules to safeguard their logical and semantic validity. Firstly, the head
of a clause \emph{must} be an object-type variable, for else the pattern is bound to a specific resource, making
generalization impossible. Secondly, for all-but-one object-type variables in the head of a clause there \emph{must} exist a
clause which has the same variable in the tail position, thus ensuring a connected graph pattern. Third and
final, the tail of a non-terminal clause \emph{must} be an object-type variable: ending such as clause with a data-type
variable, a value-range variable, or a literal is semantically invalid, whereas ending it with an entity is nonsensical
since any continuation from that point onwards will not result in a reduction of the pattern's domain. In contrast, the
tail of a \emph{terminal} clause can be a resource or any kind of variable.

We organize graph patterns based on depth, length, width, and support. The depth of a pattern equals the longest path
between any two elements, whereas the length and width equal the number of clauses in total and the maximum number of
clauses with the same head, respectively. The support value equals the number of occurrences of a pattern in a
specific dataset. An example graph pattern is depicted in Figure~\ref{fig:example_graph}-right, which has a depth of
four hops, a length and width of five and three clauses, respectively, and with an unknown support value. The logical
equivalent of this pattern is listed in Listing~1-left.

\begin{figure}[t]
\begin{minipage}[t]{0.30\linewidth}
\centering
{\footnotesize
\begin{align*}
    \phi =~ &\textit{has\_gender}(v_i, \texttt{Female})\\
    \land~  &\textit{has\_occupation}(v_i, \texttt{H.0-2}) \\
    \land~  &\textit{has\_subject}(v_j, v_i)\\
    \land~  &\textit{has\_type}(v_j, \texttt{Death\_Certificate})\\
    \land~  &\textit{at\_age}(v_j, \mathcal{N}(24.5, 1.32))
\end{align*}}
\end{minipage}
\hspace{0.5cm}
\begin{minipage}[t]{0.60\linewidth}
\centering
\begin{lstlisting}
SELECT ?$v_i$ ?$v_j$
WHERE {  
    ?$v_i$ has_gender Female .
    ?$v_i$ has_occupation H.0-2 .

    ?$v_j$ has_subject ?$v_i$ .
    ?$v_j$ has_type Death_Certificate .
    ?$v_j$ at_age ?$v_k$ .

    FILTER (
           ?$v_k$ >= "23.35"^^int
        && ?$v_k$ <= "25.65"^^int
    )
}
\end{lstlisting}
\end{minipage}
{\label{lst:example_pattern}
    \textbf{Listing 1}: The graph pattern from Fig.~\ref{fig:example_graph}-right in logical notation (left) and as SPARQL
    query (right). Variables $v_i$ and $v_j$ correspond to the two unbound resources in the figure. Note that, for brevity, the
    namespaces have been omitted.
}
\end{figure}

\section{Discovering Graph Patterns}

Our algorithm employs a two-phase approach for discovering graph patterns. During the first phase, the algorithm generates
all possible single-clause patterns that satisfy the minimal requested support. These so-called \textit{base patterns}
form the building blocks for more complex graph patterns, which are generated during the second phase by extending previously
discovered graph patterns with appropriate base patterns. Since all complex graph patterns are a combination of base
patterns, and since generating and evaluating new patterns involve simple set operations, minimal further resource-intensive
computation is necessary after completing the first phase. By also providing each pattern with a description of its
domain (e.g.\ via a set of integer-encoded resources) we no longer require to keep the original graph in memory while
retaining the minimal information necessary to derive the domain and support for new patterns.

New graph patterns are generated \textit{breadth first}, by first computing all possible patterns of minimal size and by
then iteratively combining these to form ever more complex patterns. This gives our algorithm the \textit{anytime} property,
as users can terminate a run at their leisure while still obtaining potentially-interesting, albeit less complex,
results. Our algorithm is also \textit{embarrassingly parallel}, as each new pattern effectively starts a separate
branch which can be computed independent from any of the other branches.

Please note that, for the purpose of conciseness, all procedures shown are simplified by leaving out pruning points and other
optimization techniques.

\subsection{Constructing Base Patterns}

\begin{algorithm*}[t]
    \floatname{algorithm}{Procedure}
    \caption{The procedure (simplified) for computing all base patterns with a minimal support value. Only the case
        with a single object-type variable ($\upsilon_{ot}^t$) is shown (line 12); the other cases, which have
    variables on both sides of the clause, are similar but require an extra step to calculate the domain and/or
range.}\label{alg:basepatterns}
    \begin{algorithmic}[1]
        \Function {ComputeBasePatterns}{$G$, $supp_{min}$}
            \State $\Omega := $ empty list
            \State $\Omega.addItem(\text{empty map})$
            \For{type $t$ in $\{t \mid \exists e \in \mathcal{E}: type(e, t)\}$}
                \State $\mathcal{B} := $ empty set
                \State $\mathcal{E}_t := \{e \in \mathcal{E} \mid type(e, t)\}$
                \If {$|\mathcal{E}_t| \ge supp_{min}$}
                    \For{$p \in \mathcal{P}$}
                        \State $\mathcal{U} := \{p(e, r) \mid \exists e \in \mathcal{E}_t, \exists r \in \mathcal{R}: p(e, r) \in \mathcal{A})\}$
                        \For {$ p(\cdot, r) \in \mathcal{U}$}
`                               \If {$|p(\cdot, r) \in \mathcal{U}| \ge supp_{min} $}
                                \State $\phi := p(\upsilon_{ot}^t, r)$
                                \State $\mathcal{B} := \mathcal{B} \cup \{\phi\}$
                            \EndIf
                        \EndFor
                    \EndFor
                \EndIf
                \State $\Omega(0, t) := \mathcal{B}$
            \EndFor
            \State \textbf{return} $\Omega$
        \EndFunction
    \end{algorithmic}
\end{algorithm*}

Base patterns are generated by generalizing over all assertions of which the entity in the head position is of the same type, as
shown in Procedure~\ref{alg:basepatterns}. This type-centric approach is chosen because the members of a class are
likely to possess similar characteristics and, by extension, are also likely to share similar regularities. By replacing the
specific head entities in these assertions by their corresponding object-type variables, we obtain clauses of
the form $p(\upsilon_{ot}^t, r)$ which represent a relationship $p$ between an entity of type $t$ and a resource $r$.
After computing the domain and support, each clause that enjoys a sufficiently high score is made into a
pattern, $\phi = p(\upsilon_{ot}^t, r)$, and added to polytree $\Omega$ as root.

For brevity, the pseudocode in Procedure~\ref{alg:basepatterns} omits the computation of clauses with a variable in the
tail position. Similar steps can be used, however, to generate the remaining three cases: for object-type and data-type
variables, we simply need to keep count of the various types of entities and literals, respectively, and, when this
count meets the minimal requested support, create a new base pattern $\phi = p(\upsilon_{ot}^t, \upsilon_{ot}^{t'})$ or
$\phi = p(\upsilon_{ot}^t, \upsilon_{dt}^{t'})$, with $\upsilon_{dt}^{t'}$ a data-type variable of type $t'$, which then
gets added to $\Omega$. For value-range variables, however, a few additional steps are needed.

\begin{algorithm}[t]
    \floatname{algorithm}{Procedure}
    \caption{The procedure (simplified) to iteratively discover more complex graph patterns, by matching possible endpoints (line 9) with appropriate base patterns (line 11). Function $\Delta(\cdot)$ returns the depth of an element.}\label{alg:main}
    \begin{algorithmic}[1]
        \Function {Discover}{$G$, $supp_{min}$, $d_{max}$}
        \State $\Omega :=~  $\texttt{ComputeBasePatterns}$(G, supp_{min})$
            \Statex
            \State $d := 0$
            \While{$d < d_{max}$}
                \For{type $t$ in $\Omega.types()$}
                \State $\mathcal{O} := $ empty set
                   \For{$\phi \in \Omega(d, t)$}
                        \State $\mathcal{C} := $  empty set
                        \State $\mathcal{I} := \{c = p(\cdot, \upsilon_{ot}^{t}) \mid c \in \phi \land \Delta(\upsilon_{ot}^{t}) = d \}$
                        \For{$c_i = p_k(\cdot, \upsilon_{ot}^{t}) \in \mathcal{I}$}
                            \State $\mathcal{J} := \{c = p(\upsilon_{ot}^{t}, \cdot) \mid c \in \Omega(0, t)\}$
                            \For{$c_j = p_l(\upsilon_{ot}^{t}, \cdot) \in \mathcal{J}$}
                                \State $\mathcal{C} := \mathcal{C} \cup \{(c_i, c_j)\}$
                            \EndFor
                        \EndFor
                        \State $\mathcal{O} := \mathcal{O} \cup \texttt{Explore}(\phi, \mathcal{C}, supp_{min})$
                    \EndFor
                    \State $\Omega(d+1, t) := \mathcal{O}$
                \EndFor
            \State $d := d+1$
            \EndWhile
            \State \textbf{return} $\Omega$
        \EndFunction
    \end{algorithmic}
\end{algorithm}

Value-range variables are generated by defining one or more distributions over all the literal values that occur on the
right-hand side of a relationship $p$ with entities of type $t$. For numerical data, this involves fitting multiple
Gaussian mixture models with various seeds and different number of modes, and by then evaluating these fits using the
Bayesian Information Criterion (BIC). For temporal data, such as dates, months, and durations, the same procedure is followed
but now the values are first converted into seconds (Unix time). Additionally, in either case the values are standardized, shuffled,
and augmented with a tiny amount of Gaussian noise to improve the fit. The fitted distributions $F_1, F_2, \ldots, F_n$ are made into
value-range patterns $p(\upsilon_{ot}^t, \upsilon_{vr}^{F_i})$ if the number of literal values they cover meets
the minimal requested support.

The final variant of a value-range variable targets textual data, including natural language and arbitrary strings, and
involves the generation of hierarchical regular expressions. This is accomplished by first generating regular
expressions for each value separately, clustering these by similarity, and by then generalizing the expressions until they
cover (almost) all members. We align these expressions with our earlier definition of a domain by regarding them as
uniform distributions to specific character sets. 

\subsection{Combining Graph Patterns}

\begin{algorithm}[t]
    \floatname{algorithm}{Procedure}
    \caption{The produce (simplified) to evaluate the candidate extensions, and all legal combinations thereof. Function \texttt{supp($\cdot$)} returns the support value for a given pattern.}\label{alg:explore}
    \begin{algorithmic}[1]
        \Function {Explore}{$\phi$, $\mathcal{C}$, $supp_{min}$}
        \State $\mathcal{O} := $ empty set
            \State $Q := $ empty queue
            \Statex
            \State $Q.enqueue(\phi)$
            \While{$Q \neq \emptyset$}
                \State $\psi := Q.dequeue()$
                \For {$c_i, c_j \in \mathcal{C}$}
                    \State $\psi' := \psi \land c_j$\Comment $\psi = c_1 \land c_2 \land \ldots \land c_i$
                    \State \textbf{if} $\texttt{supp}(\psi') \geq supp_{min}$
                    \State \textbf{then}
                    \State \hspace{\algorithmicindent}$\mathcal{O} := \mathcal{O} \cup \{\psi'\}$
                    \State \hspace{\algorithmicindent}$Q.enqueue(\psi')$
                \EndFor
            \EndWhile
            \State \textbf{return} $\mathcal{O}$
        \EndFunction
    \end{algorithmic}
\end{algorithm}

Going from the base patterns to more complex patterns involves the generation of candidate extensions $\mathcal{C}$,
which, if deemed favourable, are appended to their parents' set of clauses to form new graph patterns $\psi'$. These new
patterns are then added to $\Omega$ as children to their parents, provided that they meet the minimal requested support.
Procedure~\ref{alg:main} and \ref{alg:explore} outline this process.

Each of the candidate extensions is a pair of clauses $(c_i, c_j)$, where $c_i = p_k(\cdot, \upsilon_{ot}^{t})$ is one
of the parent's outer clauses---a candidate endpoint---and $c_j = p_l(\upsilon_{ot}^{t}, \cdot)$ is a suitable base
pattern. Both clauses are ensured to hold the same object-type variable, thus providing a semantically valid connection. 
Depending on the element in the tail position of clause $c_j$, the extension, if added, will be terminal or
non-terminal. If a pattern has no further non-terminal clauses it will be omitted from future iterations of the
algorithm.

There are often multiple extensions possible per pattern within the same iteration. To exhaustively explore the space of
multiple extensions, our algorithm evaluates each of these extensions separately, as well as their $k$-combination
(without repetition) with $k$ ranging from two to the number of candidate extensions $|\mathcal{C}|$. Note that, since not all
combinations are accepted, the actual maximum value for $k$ will generally be less than $|\mathcal{C}|$ in practice.

\begin{figure}[t]
  \centering
    \includegraphics[width=.5\textwidth]{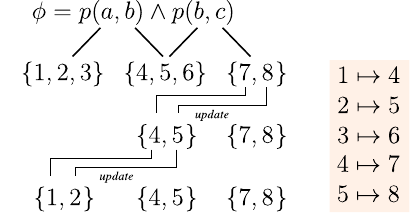}
    \caption{Updating the domain of a pattern $\phi$ after adding clause $p(b,c)$. Domains are depicted as sets with
    integer-encoded resources, whereas the maps between resources represent assertions in the graph. Since resource $6$
    is not connected to any of the resources in the domain of $c$, adding $p(b,c)$ thus reduces the domain of $b$ (by
removing resource 6), which, in turn, reduces that of $a$ (by removing resource 3).}
    \label{fig:domain_prop}
\end{figure}

For each new pattern we compute the domain and associated support score. Since patterns carry a description of their own
domain, we can easily compute the domain of a newly derived pattern by taking the intersection of its parent's domain with
that of the recently-added clause, and by then propagating this change through the other clauses in the pattern.
Figure~\ref{fig:domain_prop} illustrates this principle, by showing how adding a new clause reduces the domain of the
pattern as a whole.

\subsection{Search Optimization}

Smart pruning techniques and other optimizations are used to reduce the search space by avoiding duplicate, invalid,
and/or poorly supported patterns and clauses. We provide a brief description of the most important techniques next.

\begin{itemize}
    \item Since every added clause makes a pattern more specific, it must follow that the corresponding domain
          should be a proper subset of that of its parent. Hence, patterns that have the same domain as their parent
          are pruned and disallowed from becoming a parent themselves. The sole exception are clauses with an
          object-type variable as tail, which are kept for one iteration more in case they might farther a pattern that
          \emph{does} reduce the domain.

      \item Patterns that were not extended during the current iteration are omitted from future iterations. The
          intuition behind this is that future iterations necessarily involve more specific patterns; if the patterns
          did not meet the minimal support during the current iteration, then it follows that this will also be the case
          for future iterations. The same holds for base patterns.

      \item Candidate extensions that do not meet the minimal required support are omitted from future iterations. Since
          the domain of an extension will stay unchanged during the entirety of a run, it follows that adding them can
          never result in a pattern with a sufficiently high support score. Base patterns for which this is the case are
          already filtered during their creation.

      \item Duplicate patterns (which might occur via different routes) are pruned early on by creating
            a cheap proxy---the logical formula as string---and checking this against a hash table before creating the
            actual object and computing its domain.

        \item Patterns that only have terminal clauses or no appropriate object-type variables are disallowed from
            becoming a parent, whereas patterns which exceed the maximum allowed length, width, or depth are pruned
            early on for obvious reasons. 
\end{itemize}

\subsection{Pattern Browser}

\begin{figure}[t]
  \centering
  \includegraphics[width=\textwidth]{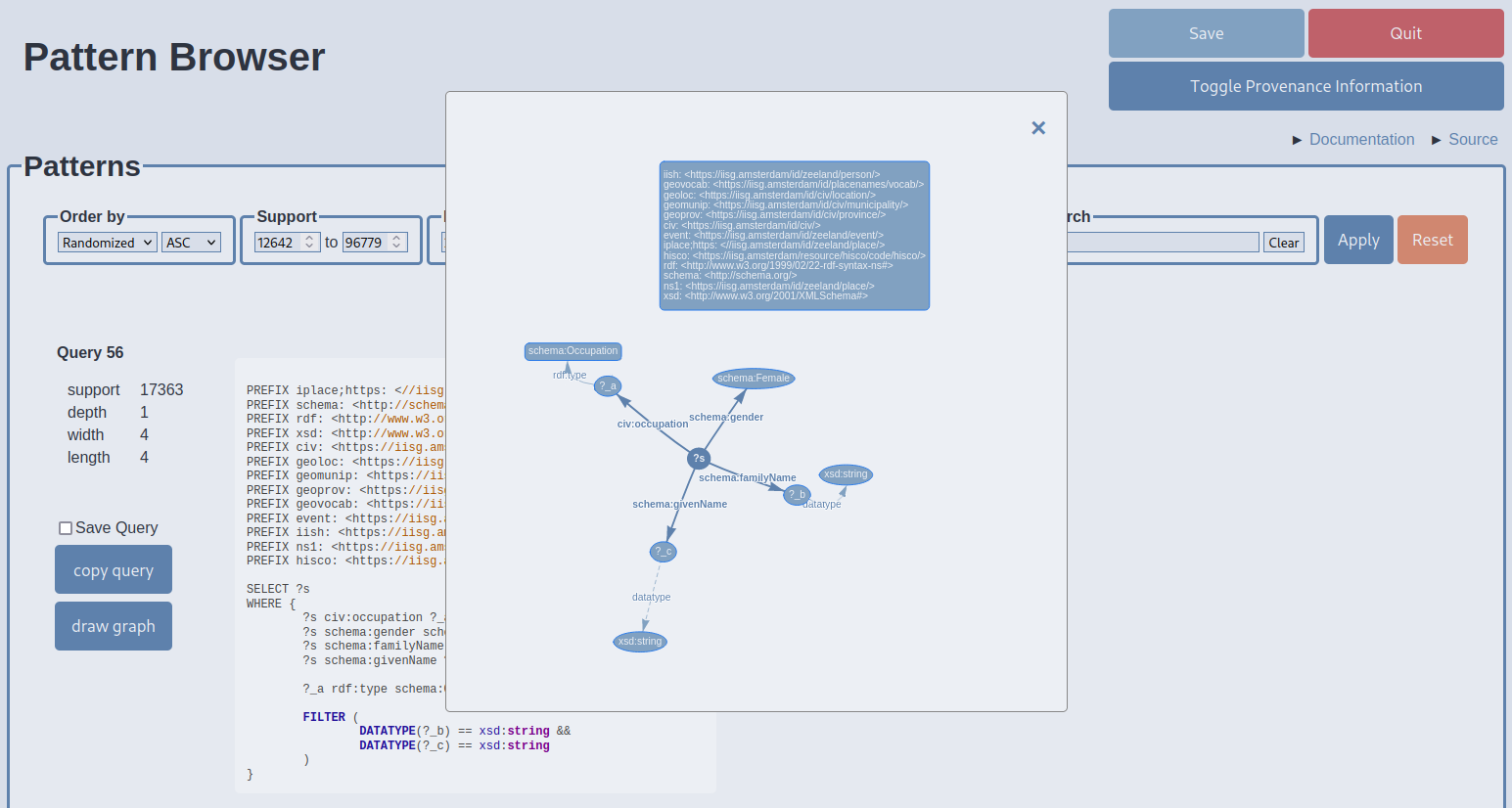}
  \caption{A screenshot of the facet browser showing a graph pattern encoded as SPARQL query and visualized as a graph.}
    \label{img:browser}
\end{figure}

A simple facet browser (Fig.~\ref{img:browser}) was created to assist scholars with the exploration and analysis of the
discovered patterns. Build upon open web standards, the pattern browser facilitates the filtering of patterns over various
dimensions, including \textit{support}, \textit{depth}, \textit{length}, and \textit{width}, as well as provide
full-text search capabilities. The filtered selection can be saved in a separate file, which itself can be opened in the
pattern browser for further analysis.  Alternatively, the saved selection can be shared with others or published on the
web, facilitation reuse and reproducibility.

Metadata is stored together with the patterns and can be viewed directly from the pattern browser. Amongst others, these
data include provenance information and hyperparameter settings from the process that created the patterns. Additional
data is appended to the provenance information upon saving a filtered selection, allowing users to trace back all
performed actions. Both patterns and metadata are stored using RDF.

\section{Evaluation}

We evaluate our algorithm and the patterns it produces from a user-centric perspective, by conducting a user study
amongst a select group of domain experts from the humanities. The primary goal of this user study is to ascertain the
perceived interestingness of the discovered patterns, as well their interpretability. For this purpose, graph patterns
were discovered within a domain-specific knowledge graph and presented to experts to assess. The implementation of
our algorithm that was used to generate these patterns is available online\footnotemark. All runs of
this algorithm were performed on the DAS-6 supercomputer~\cite{DBLP:journals/computer/BalELNRSSW16}. 

\begin{figure}[t]
\begin{minipage}[t]{0.45\linewidth}
\centering
\begin{lstlisting}
SELECT ?$v_i$ ?$v_j$
WHERE {  
    ?$v_i$ has_gender Male .
    ?$v_i$ has_occupation H.61220 .
    ?$v_i$ has_age ?$v_k$ .

    ?$v_j$ has_subject ?$v_i$ .
    ?$v_j$ has_type Mariage_Certificate .

    FILTER (
      ?$v_k$ == "29"^^int
    )
}
\end{lstlisting}

\fbox{%
\parbox{0.90\linewidth}{%
"In this population sample, 1,238 out of 100,000 records are about 29 year old married men who work in agriculture."
 }%
}%

\end{minipage}
\hfill
\begin{minipage}[t]{0.56\linewidth}
\centering
\begin{lstlisting}
SELECT ?$v_i$
WHERE {  
    ?$v_i$ has_gender Female .
    ?$v_i$ has_firstName ?$v_j$ .
    ?$v_i$ has_familyName ?$v_k$ .

    FILTER (
      REGEX(?$v_j$, "[a-z]{2,14}\s[a-z]{2,14}")
      && REGEX(?$v_k$, "[a-z]{3,16}")
    )
}
\end{lstlisting}

\vspace{2.4mm}
\fbox{%
\parbox{0.90\linewidth}{%
"In this population sample, 13,632 out of 100,000 records are about women with two first names between 2 and 14
characters each, and with a family name between 3 and 16 characters."
 }%
}%

\end{minipage}\vspace{5mm}
{\label{lst:discovered_patterns}
    \textbf{Listing 2}: Two graph patterns that were discovered in the civil registry dataset, encoded as SPARQL
    queries, together with their natural language description. Note that, for brevity, the
    namespaces have been omitted.
}
\end{figure}

\footnotetext{See \url{gitlab.com/wxwilcke/hypodisc}}

\subsection{Dataset}

The knowledge graph used in our experiments contains the civil records from Dutch citizen who were alive between 1811
and 1974. Each record includes information about a person's pedigree, marital status, occupation, and location, as well
as various important life events including birth, death, and becoming a parent. Due to the sensitivity of these data
we are prohibited from sharing this dataset, unfortunately.

In its entirety, the dataset contains the records from over 5.5 million people. For experimental purposes, a subset was
created by randomly sampling 100 thousand individuals together with their context, resulting in a graph with just over
one million assertions between roughly 635 thousand resources. We believe that these numbers are sufficiently large enough for the
same patterns to emerge as those present in the original dataset. 

Two example patterns that were found during these experiments are listed in Listing 2, together with their description
in natural language. The left-hand pattern covers the set of all 29 year old men who are married and who work in the
agricultural sector, which accounts for 1,238 individuals in the dataset. The graph pattern on the right accounts for
13,632 people, and encompasses all women with two first names between two and 14 characters each, and with a
family name between three and 16 characters.

\subsection{User Study}

The user study took the form of an online
questionnaire, lowering the barrier for participation and allowing for a cross-border audience. To ensure a good fit
between this audience and the topic at hand it was decided to make the questionnaire open to invitation only. While this
might have resulted in a lower number of participants, we are confident that their responses are more valuable. 

The questionnaire was split into four sections. In the first section, participants were asked about their familiarity
with the core concepts surrounding this research. The answers to these questions allowed us to weight the participants'
responses on later questions. The second and third sections involved questions about the graph patterns and the pattern
browser, respectively, whereas the last section asked several overarching questions about the perceived usefulness of
our method and the patterns it yields. In all cases (save for open questions) the responses were recorded using a
five-point Likert scale ranging from \textit{Strongly Disagree} (negative) to \textit{Strongly Agree} (positive).

To assess the graph patterns on interestingness~\cite{DBLP:series/sci/GengH07}, participants were presented
with several hand-picked patterns in SPARQL format, and asked to rate each one on \textit{novelty},
\textit{validity}, and \textit{utility}, as well as on \textit{interpretability}. A similar setup was used for
the pattern browser, but instead using screenshots and rated on \textit{helpfulness} (in analysing
the patterns), \textit{intuitiveness} (of the interface), \textit{pleasantness} (of the colour scheme), and
\textit{understandability} (of the displayed information).

To determine the agreement and reliability amongst participants, we employ Kendall's coefficient of concordance $W$
for its suitability to evaluate ordinal data with multiple ratings over multiple items~\cite{kendall1948rank}. For
similar reasons, we use Kendall rank correlation coefficient $\tau$ to measure dependencies between
responses~\cite{kendall1938new}. We furthermore employ factor analyses to obtain a better understanding of the
interactions between criteria.

\begin{figure}[t]
  \centering
    \includegraphics[width=\textwidth]{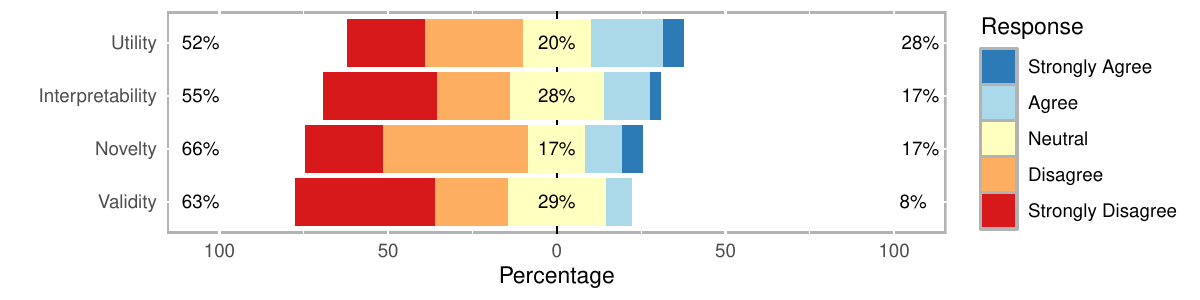}
    \caption{Responses on a 5-point Likert scale (Kendall's $W = 0.14$) about the perceived novelty, validity,
        utility, and interpretability of the presented graph patterns.}
    \label{fig:likert_patterns}
\end{figure}

\begin{figure}[t]
  \centering
    \includegraphics[width=\textwidth]{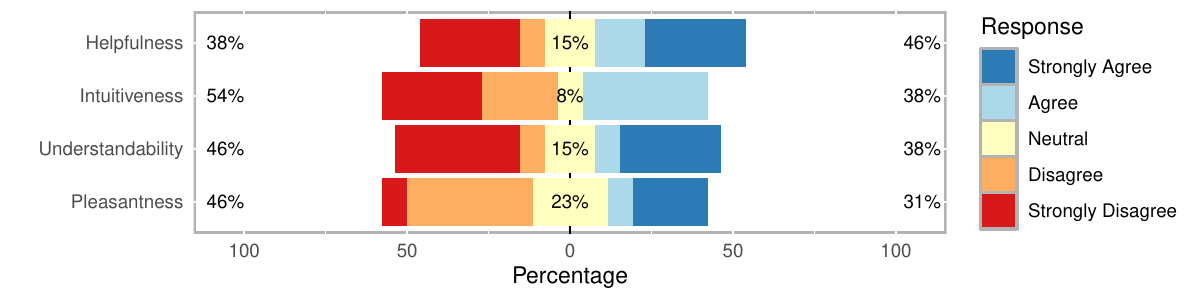}
    \caption{Responses on a 5-point Likert scale (Kendall's $W = 0.11$) about the perceived helpfulness, intuitiveness,
        understandability, and pleasantness of the pattern browser.}
    \label{fig:likert_browser}
\end{figure}

\begin{table}[t]
   \parbox{.63\linewidth}{
    \centering
    \caption{Familiarity of the participants (Kendall's $W = 0.29$) with the domain, knowledge graphs, SPARQL, and
    database terminology. Last column shows correlation (Kendall's $\tau$) with
perceived utility (Tab.~\ref{tab:usefulness}).}
    \label{tab:familiar}
    \begin{tabular}{p{1.9cm}p{2.2cm}p{2.5cm}r}
        \toprule
        Familiarity & Median & Mode & $\tau$ \\\midrule
        Domain & \textit{neutral} & \textit{neutral} & -0.37\\
        Graphs & \textit{disagree} & \textit{neutral} & 0.42\\
        SPARQL & \textit{disagree} & \textit{strongly disagree} & 0.51\\
        DB Terms & \textit{disagree} & \textit{disagree} & 0.77\\
        \bottomrule
    \end{tabular}
        }\hfill
    \parbox{.33\linewidth}{
    \centering
    \caption{Utility of the patterns (Kendall's $W = 0.54$) as perceived by participants in relative numbers.}.
    \label{tab:usefulness}
    \begin{tabular}{lc}
        \toprule
        Score & Portion\\\midrule
        \textit{fully agree} & 0.00 \\        
        \textit{agree} & 0.15 \\
        \textit{neutral} & 0.24 \\
        \textit{disagree} & 0.15 \\
        \textit{fully disagree} & 0.46\\
        \bottomrule
    \end{tabular}
    }
\end{table}

\subsection{Results \& Discussion}

A total of 13 out of the 42 experts on social and economic history to whom we reached out took part in the user study,
corresponding to a fair response rate of 31\%. Table~\ref{tab:familiar} lists their
familiarity with the domain, knowledge graphs, SPARQL, and database terminology. The responses suggest that the
participants only moderately align with the domain, as both the median and mode scores are \textit{neutral}. Similar for
their familiarity with knowledge graphs, albeit with a lower median of \textit{disagree}. Even less familiar do the
participants seem to be with SPARQL and database terminology, having a median of \textit{disagree} and mode of
\textit{strongly disagree} for the former, and a median and mode of \textit{disagree} for the latter. Together, these
responses suggest a possible gap between the technical background and experience of the participants and the skills
required to fully comprehend the method and the patterns it yields. While this discrepancy might not be evenly spread
amongst the participants, as suggested by the relatively low agreement ($W = 0.29$), it may have induced a degree of
uncertainty in the participants and in the answers they have provided. This
is further corroborated by the self-reported confidence (Table~\ref{tab:confidence}), with roughly half of the participants
(46\%) giving themselves the lowest score ($W = 0.85$).

Figure~\ref{fig:likert_patterns} shows the responses about the presented graph patterns. Overall, the provided scores
suggest that the participants were critical about the discovered patterns, with 52\% to 64\% believing the patterns to
be uninteresting against 8\% to 28\% deeming the opposite. However, the number of people who were very negative differ
considerably from roughly one out of three to two out of three negative responses. This is supported by the low
agreement ($W = 0.14$) amongst raters, which indicates a wide range of opinions. Looking at the underlying criteria, we
observe that the participants seem the most positive (28\%) about the utility of the patterns, followed by their novelty
(17\%). The patterns' validity, however, scores poorly with only few participants being positive (8\%). This last score
is particularly interesting since the patterns are generalisations of the original data, rather than predictions, and
are therefore as valid as the data they are discovered on. That the patterns were nevertheless deemed invalid by most of
the participants suggests that there are either problems with the chosen dataset (which is unlikely, it being a curated
dataset) or that there was a mismatch between the experts' expectations and the output of our method. This latter reason
seems more probable, since only few participants were positive about interpretability (17\%).  

An analysis of the factor loadings belonging to these responses (Table~\ref{tab:fa_patterns}) shows a clear separation
between criteria, with utility ($0.90\lambda_1$) and validity ($0.88\lambda_1$) on one hand, and interpretability
($0.97\lambda_2$) on the other. This suggests that both utility and validity contribute to the same latent component,
which we can perhaps interpret as an indication of \textit{effectiveness}, whereas interpretability measures an entirely
different component of its own. Less clear cut is novelty, which enjoys significant cross loadings on both components
($-0.54\lambda_1 + 0.40\lambda_2$) which suggests that this criterion is a poor indicator for the dimensions on which
the participants assess the usefulness of the patterns. Rather, novelty appears to be a combination of low effectiveness
and high interpretability, suggesting that it is a product of our method as opposed to an inherent characteristic. This
creates a peculiar paradox, where users rate the effectiveness based on the method's ability to discover known and useful
patterns, but value novel insights for their perceived validity and utility as long as they are easy to understand.

Participants were largely divided about the pattern browser (Figure~\ref{fig:likert_browser}), with 31\% to 46\% seeing the
tool as beneficial and user friendly against 38\% to 46\% thinking otherwise. This large range is again supported by the
low agreement between participants ($W = 0.11$). In terms of helpfulness and understandability the number of (very) positive
reactions are largely in balance with the (very) negative reactions; the helpfulness scores the most positive with
almost half of the participants (46\%) deeming the browser beneficial for analysing patterns, while a large portion
(38\%) of participants is also relatively positive about how the browser conveys the patterns in an way that is
understandable. Respondents were more critical of the browser's intuitiveness and pleasantness. While a comparable number of
participants assessed the intuitiveness as either positive or negative, there were none who were very positive.
Conversely, only few negative respondents were very negative (8\%) about the pleasantness of the colour scheme used by the
interface, whereas most positive participants were very positive (23\%).

\begin{table}[t]
   \parbox{.45\linewidth}{
    \centering
    \caption{Factor loadings of the responses on the presented graph patterns, averaged over participants, using an
oblique rotation (\textit{BentlerQ}\cite{bentler1977factor}) with two components which, together, account for 87\% of the total variation.}
    \label{tab:fa_patterns}
    \begin{tabular}{p{3cm}rr}
        \toprule
        Criterion & $\lambda_1$ & $\lambda_2$ \\\midrule
        Novelty & -0.54 & 0.40 \\
        Validity & 0.88 & -0.05 \\
        Utility & 0.90 & 0.14 \\
        Interpretability & 0.03 & 0.97 \\
        \bottomrule
    \end{tabular}
        }\hfill
    \parbox{.45\linewidth}{
    \centering
    \caption{Factor loadings of the responses on the pattern browser, using an oblique rotation
        (\textit{Simplimax}\cite{kiers1994simplimax})
    with two components which, together, account for 86\% of the total variation.}
    \label{tab:fa_ui}
    \vspace{4mm}
    \begin{tabular}{p{3cm}rr}
        \toprule
        Criterion & $\lambda_1$ & $\lambda_2$ \\\midrule
        Intuitiveness & 0.61 & 0.29 \\
        Pleasantness & 0.02 & 0.99 \\
        Helpfulness & 1.00 & -0.16 \\
        Understandability & 0.85 & -0.02 \\
        \bottomrule
    \end{tabular}
    }
\end{table}

The factor loadings that belong to these responses are listed in Table~\ref{tab:fa_ui}, and indicate a strong divide
between pleasantness ($0.99\lambda_2$) and the other three criteria: intuitiveness ($0.61\lambda_1$), helpfulness
($1.00\lambda_1$), and understandability ($0.85\lambda_1$). A likely explanation is that pleasantness measures purely
the visual appearance of the browser, whereas the remaining three are a measure of the browser's usefulness.
Intuitiveness stands out, however, by also providing a moderate contribution ($0.29\lambda_2$) to the visual component.
This might be explained by that this criterion, like novelty, is a product of the other dimensions rather than an
inherent characteristic, suggesting that intuitiveness stems from whether users deem the browser intelligible and easy
to use.

Table~\ref{tab:usefulness} lists the overall utility of the graph patterns and browser as perceived by the participants,
and suggests an overall critical opinion with 15\% of the experts agreeing that the patterns and/or browser can
be useful. Different from the other responses, this opinion enjoys a much higher, albeit still moderate, agreement
($W=0.54$). Correlation tests with the participants' familiarity scores show a substantial positive correlation ($\tau =
0.77$) between having a strong negative opinion and having little experience with database terminology, and moderate
positive correlation with the unfamiliarity with SPARQL ($\tau = 0.51$) and knowledge graphs ($\tau = 0.42$). This
suggests that scholars who possess a more inductive, data-focussed, mindset were more positive about our
approach, whereas more deductive, theory-minded, scholars were most critical.

Remarks left by the experts shed some light on the results. While a variety of reasons were given, the large majority of
these can be summarized as "missing the context". According to these experts, it is difficult to infer anything useful
from the patterns if presented in isolation. Rather, more detailed information should be provided on the data and the
domain they cover. Other insight that can be gained from the remarks is the strong preference for a natural language
representation, rather than the SPARQL format or graph visualization, despite the likely loss of precision due to the
translation. A final common remark is the degree of interestingness, which still varies too much.

\begin{table}
    \centering
    \caption{Confidence of the participants ($W = 0.85$) in relative numbers.}
    \label{tab:confidence}
    \begin{tabular}{lc}
        \toprule
        Score & Portion\\\midrule
        \textit{fully agree} & 0.00 \\        
        \textit{agree} & 0.00 \\
        \textit{neutral} & 0.38 \\
        \textit{disagree} & 0.15 \\
        \textit{fully disagree} & 0.46 \\
        \bottomrule
    \end{tabular}
\end{table}

\section{Conclusion \& Future Work}

This work introduced an \textit{anytime}, bottom-up, and easily parallelizable algorithm to efficiently discover
\textit{generalised multimodal graph patterns} in knowledge graphs. To facilitate further filtering and analysis, the
discovered patterns are converted to SPARQL queries and presented in a simple facet browser. An evaluation of the
patterns and the browser was held in the form of a user study amongst a select group of domain expert. While reactions
were mixed, further analysis suggested that the most critical experts acted from a feeling of uncertainty caused by
their unfamiliarity with the technical skills required to fully comprehend the patterns and the method that generated
them. Rather, this group expressed their preference for more context and natural language explanations, finding it
challenging to interpret the patterns otherwise. Conversely, the experts who did posses appropriate technical
backgrounds were more positive in general, particularly where utility is concerned. 

Further analysis also revealed a peculiar, yet interesting, paradox that suggests that many experts set out to find
interesting new patterns, yet rated novel patterns more negatively because they do not conform to the current scholarly
literature or the experts' own beliefs. This effect might be a form of confirmation bias or simply a distrust of new
technologies, yet poses an intriguing conundrum since the most straightforward solution (emphasizing existing knowledge)
would invalidate the method's entire reason for being. On the other hand, since the perception of novelty appears to
emerge from other characteristics rather than being an intrinsic characteristic of a pattern itself, as
suggested by our findings, it can be argued that the primary goal should not be about finding \emph{novel} patterns, but
rather about discovering explainable connections between patterns that are already known and validated. Developing such
methods would be an interesting exercise for future work.

There are several other natural directions to follow up on in future work. First and foremost is the improvement of the
measure of interestingness, and how to steer away from uninteresting patterns. This is a common and difficult problem
with pattern mining which is largely the result of an algorithm's reliance on statistics. Expanding the method's ability
to exploit background information might help counter this by making more informed decisions when exploring the search
space, for example by favouring patterns that contain elements from a domain-specific taxonomy. Another possible
solution to avoid uninteresting patterns might be to more actively involve the users in the discovery process, by asking
them to score candidate patterns as they are discovered. This would enable scholars to fine-tuning the output to their
own expectations, further increasing explainability and transparency. These scholars can be supported by a meta
model that learns to differentiate between patterns that are interesting and those which are not, and which, once
satisfactory, can be shared with fellow researchers.

Future work might also consider further improving how scholars can inspect and analyse the discovered patterns, for
example by developing an interactive dashboard which provides detailed information about the context on various levels
of granularity. This information could include general statistics about the relevant classes and predicates, as well as
provide an overview of their semantics, their members, and other closely related elements. To increase interpretability,
the patterns themselves can perhaps be offered as natural language explanations, which could be generated automatically by
leveraging the annotations in the graph if provided, or by employing a large language model trained on similar
data.

\printbibliography

\end{document}